\documentclass[twoside]{article}

\usepackage[accepted]{aistats2017final}

\usepackage{algorithm}
\usepackage{algorithmic}
\usepackage{mathtools}
\usepackage{graphicx}
\usepackage{float}
\usepackage{caption}
\usepackage{natbib}
\setcitestyle{authoryear,open={(},close={)}}
\usepackage{hyperref}
\usepackage{amsmath,amsthm,amssymb}
\usepackage{bm}

\newtheorem{lemma}{Lemma}[]
\newcommand{\Enc}{\texttt{Enc}}
\newcommand{\Dec}{\texttt{Dec}}

\newcommand{\sml}[1]{\vcenter{\hbox{\scalebox{0.55}{$#1$}}}}
\newcommand{\sbl}{\vcenter{\hbox{\scalebox{0.45}{$\bullet$}}}}

\newcommand\degr{\text{deg}}
\newcommand{\B}{\bm{\beta}}

\newcommand{\y}{\bm{y}}
\newcommand{\X}{\bm{X}}
\newcommand{\XT}{\bm{X}^{T}}
\newcommand{\Xj}{\bm{X}_{\sbl j}}

\newcommand{\Xjj}[1]{\bm{X}_{\sbl #1}}
\newcommand{\XjjT}[1]{\bm{X}_{\sbl #1}^{T}}

\newcommand{\I}[1]{\bm{I}_{\sml{#1}}}
\newcommand{\0}[1]{\bm{0}_{\sml{#1}}}

\usepackage{enumitem}

\defcitealias{ObamaReport14}{{United States EOP, 2014}} 

\begin{document}

%

%

\twocolumn[

\aistatstitle{Encrypted accelerated least squares regression}
\aistatsauthor{
Pedro M. Esperan\c{c}a
\And
Louis J. M. Aslett
\And
Chris C. Holmes }
\aistatsaddress{
Department of Statistics \\
University of Oxford
\And
Department of Statistics \\
University of Oxford
\And
Department of Statistics \\
University of Oxford } ]

\begin{abstract}
Information that is stored in an encrypted format is, by definition, usually not amenable to statistical analysis or machine learning methods. In this paper we present detailed analysis of coordinate and accelerated gradient descent algorithms which are capable of fitting least squares and penalised ridge regression models, using data encrypted under a fully homomorphic encryption scheme.  Gradient descent is shown to dominate in terms of encrypted computational speed, and theoretical results are proven to give parameter bounds which ensure correctness of decryption.  The characteristics of encrypted computation are empirically shown to favour a non-standard acceleration technique.  This demonstrates the possibility of approximating conventional statistical regression methods using encrypted data without compromising privacy.
\end{abstract}

\section{Introduction}

Issues surrounding data security and privacy of personal information are of growing concern to the public, governments and commercial sectors. Privacy concerns can erode confidence in the ability of organisations to store  data securely, with a consequence that individuals may be reticent to contribute their personal information to scientific studies or to commercial organisations (\citealp{SandersonETAL15,NaveedETAL15,GymrekETAL13}).

In this paper we demonstrate that statistical regression methods can be applied directly to encrypted data without compromising privacy.
Our work involves adapting existing methodology for regression analysis in such a way as to enable full computation within the mathematical and computational constraints of recently developed fully homomorphic encryption schemes \citep{Gentry10,FanVercauteren12}.  We show empirically that traditional state-of-the-art convergence acceleration techniques can under-perform when such constraints are taken into account.

Fully homomorphic encryption (FHE) differs from differential privacy (DP) in that it provides for exact computation with cryptographically strong privacy of all the data during the fitting process itself, at the expense of much greater restrictions on the operations and computational cost \citep{Gentry10,AEH15review}.  However, it is complementary to DP (as opposed to competing with it) in the sense that FHE maintains privacy \emph{during} model fitting and prediction, while DP can ensure privacy post-processing of the data if the model itself is to be made public.  For an overview of DP see  \citet{DworkRoth14}.

FHE allows for secure operations to be performed on data, and statistical analysis and machine learning is the major reason why people want to perform mathematical operations on data. Thus, there is a real opportunity for machine learning scientists to be involved in shaping the research agenda in FHE (\citetalias{ObamaReport14}).
The applications of encrypted statistics and machine learning include general purpose cloud computing when privacy concerns exist, and are especially important in e-health and clinical decision support \citep{BasilakisETAL15,McLarenETAL16}.

\S \ref{sec:background} is a brief accessible introduction to FHE and \S \ref{sec:LR_LR} recaps regression to fix notation and our method of representing data prior to encryption.  A detailed examination of coordinate and gradient descent methods in an encrypted context follows in \S \ref{sec:LR_encLR}, including encrypted scaling for correctness, computational considerations, prediction, inference, regularisation and theoretical proofs for parameters in a popular FHE scheme.  \S \ref{sec:LR_accelerate} discusses acceleration methods optimal for encrypted computation, with examples provided in \S \ref{sec:LR_results} and discussion in \S\ref{sec:LR_discussion}.

\section{Homomorphic encryption}
\label{sec:background}

A common technique for ensuring the privacy of data is to encrypt it, but typically once one wishes to fit a model it is necessary to first decrypt and risk exposing the data.  However, recent advances in cryptography enable a very limited amount of computation to be performed directly on the encrypted content, rendering the correct result upon decryption.

\subsection{Standard cryptography}

A public key encryption scheme is one which has two algorithms, $\Enc(\cdot, \cdot)$ and $\Dec(\cdot, \cdot)$, to perform encryption and decryption respectively, together with two keys: the public key, $k_p$, can be widely distributed and used by anyone to encrypt a message; the secret key, $k_s$, is required to decrypt any message encrypted using $k_p$ and so is kept private.  The fundamental identity is:
\[ \Dec(k_s, \Enc(k_p, m)) = m \quad \forall \, m.  \]
The data to be encrypted, $m$, is referred to as the \emph{message} or \emph{plaintext} and after encryption is referred to as the \emph{ciphertext}.  In conventional encryption algorithms manipulation of the ciphertext does not typically lead to meaningful modification of the message.

\subsection{Fully Homomorphic encryption}
\label{sec:HE}

An encryption scheme is said to be \emph{fully homomorphic} if it also possesses two operations, $\oplus$ and $\otimes$, which satisfy the following identities:
\begin{align*}
  \Dec(k_s, \Enc(k_p, m_1) \oplus \Enc(k_p, m_2)) &= m_1 + m_2 \\
  \Dec(k_s, \Enc(k_p, m_1) \otimes \Enc(k_p, m_2)) &= m_1 \times m_2
\end{align*}
for all $m_1, m_2$, which can be applied a theoretically arbitrary number of times.  In other words, a homomorphic encryption scheme allows computation directly on ciphertexts which will correctly decrypt the result as if the corresponding operations had been applied to the original messages.

However there are many constraints in practical implementation, reviewed in \cite{AEH15review}.  For the purposes of this work they may be synopsised as:
\setlist{noitemsep}
\begin{itemize}
  \item Typically $m$ can only easily represent binary or integers.
  \item Data size grows substantially after encryption.
  \item The computational cost of $\oplus$ and $\otimes$ is orders of magnitude higher than standard $+$ and $\times$.
  \item Operations such as comparisons ($==, <, >$) and division ($\div$) are not possible.
  \item Implementation of current schemes necessitate a highly computationally expensive `bootstrap' operation (unrelated to statistical bootstrap) which must be applied frequently between $\otimes$ operations to control the noise in the ciphertext.
\end{itemize}

Consequently, a crucial property of any FHE scheme is the \textit{multiplicative depth}\footnote{Having only addition and multiplication operations mean all computations form polynomials.  Simply put, the multiplicative depth corresponds to the degree of the maximal degree term of the polynomial minus one.  e.g., $\sum_{j=1}^{P} a_{j}b_{j}$ has depth 1 and $\prod_{j=1}^{P} a_{j}$ has depth $P-1$.} which is possible before a `bootstrap' is required.  Typically, one can select the parameters of the FHE scheme to support a pre-specified multiplicative depth, but this is a trade-off because parameters supporting greater depth between `bootstraps' also result in larger ciphertexts and slower homomorphic operations ($\oplus, \otimes$).  Therefore, it is essential to consider the \textit{Maximum Multiplicative Depth} (MMD) required to evaluate an algorithm encrypted, since this dramatically affects speed and memory usage.
Indeed, cryptographic parameters are typically chosen to match the MMD of the target algorithm being run so as to avoid `bootstrap' altogether (which is simply deemed too computationally costly).

The above constraints often mean that standard statistical methodology cannot be applied unmodified on encrypted content.

\subsection{Privacy preserving statistics}

There has been some work towards using FHE schemes for statistics and machine learning.  Often this has involved identifying existing algorithms which can be run with minimal modifications \citep{WuHaven12,GLN13}, or fitted on unencrypted data for later prediction on new encrypted data \citep{DowlinETAL15}.
However, some recent work has also begun on developing new methodology inspired by traditional techniques, specifically tailored to homomorphic computation so that the whole analysis --- model fitting and prediction --- can be computed encrypted \citep{AEH15stats}.

In particular, the topic of linear regression in the context of FHE has not been covered systematically in the literature thus far.
\citet{HFN11} propose protocols for regression analysis which involve substantial communication and intermediate decryption between multiple parties: it takes two days to complete a problem of size $N=51,016$ observations and $P=23$ predictors.
In this work, we want to develop methods capable of fitting and prediction without any intermediate communication or decryption.
\citet{WuHaven12} were the first to tackle linear regression in the context of FHE by using Cramer's rule for matrix inversion. Unfortunately, this approach calls for the computation of the determinant of $\XT\X$, which quickly becomes intractable for even low dimensional problems (e.g., $P>4$). Specifically, the multiplicative depth is unbounded with growing dimension and, consequently, bootstrapping seems unavoidable.
\citet{GLN13} mention that regression can be done by gradient descent, but do not implement the method or give further details.

The approach we present in this work enjoys several notable properties: (i) estimation and prediction can both be performed in the encrypted domain; (ii) bootstrapping can be avoided even for moderately large problems, (iii) scales linearly with the number of predictors; (iv) and admits the option of $L_{2}$ (ridge) regularisation.

\section{The linear regression model}
\label{sec:LR_LR}

The standard linear regression model assumes
\begin{equation}\label{eq:lm}
	\y = \X\B + \bm{\varepsilon},
\quad \bm{\varepsilon}
\sim \mathcal{N}(\0{N}, \sigma^{2} \I{NN})
\end{equation}
where
$\y$ is a response vector of length $N$;
$\X$ is a design matrix of size $N \times P$;
$\B$ is a parameter vector of length $P$; and
$\bm{\varepsilon}$ is a vector of length $N$ of independent and Normally distributed errors with zero mean and constant variance $\sigma^{2}$.
Provided that $\XT\X$ is invertible, the ordinary least squares (OLS) solution to
\begin{align} \label{eq:ols}
&\min_{\B} || \y - \X\B ||_{2}^{2} \\
\label{eq:ols_sol}
\mbox{is } \quad & \hat{\B}_{\text{ols}}
= (\XT\X)^{-1} \XT\y.
\end{align}
%
%
%
Regularisation techniques trade the unbiasedness of OLS for smaller parameter variance by adding a constraint of the form
$\sum_{j=1}^{P} |\beta_{j}|^{\gamma} \leq c$,
for some $\gamma$ \citep{HTF09}.
Bounding the norm of the parameter vector imposes a penalty for complexity, resulting in shrinkage of the regression coefficients towards zero.
We focus on $L_{2}$ regularisation (ridge, henceforth RLS; \citealp{HoerlKennard70}), where $\gamma=2$.
Other options are available, although no one method seems to dominate the others \citep{ZouHastie05,Tibshirani96,Fu98}.
The standard solution to the regularised problem
\begin{align} \label{eq:rls1}
&\min_{\B} \left\{ ||\y-\X\B||_{2}^{2}
+ \alpha ||\B||_{2}^{2} \right\}\\
\label{eq:LR_rls_sol}
\mbox{is } \quad & \hat{\B}_{\text{rls}}(\alpha) = (\XT\X + \alpha \I{PP})^{-1} \XT\y
\end{align}
revealing that $\alpha$ is also key in converting ill-conditioned problems into well-conditioned ones \citep{HoerlKennard70}.



\subsection{Data representation and encoding}
\label{sec:LR_encoding}

Because FHE schemes can only naturally encrypt representations of integers, dealing with non-integer data requires a special encoding.
We use the transformation $\dot{z} \equiv \lfloor 10^{\phi} z \rceil \in \mathbb{Z}$, for $z \in \mathbb{R}$ and $\phi \in \mathbb{N}$.
Here, $\phi$ represents the desired level of accuracy or, more precisely, the number of decimal places to be retained.
By construction, \mbox{$\dot{z} \approx 10^{\phi} z$}, so that smooth relative distances between elements encoded in this way are approximately preserved.
The encoding can accommodate both discrete (integer-valued) data and an approximation to continuous (real-valued) data.
Throughout, we assume that covariates are standardised and responses centred before integer encoding and encryption.

\section{Least squares via descent methods}
\label{sec:LR_encLR}

Conventional solutions to  \eqref{eq:ols_sol} and \eqref{eq:LR_rls_sol} can be found in closed form using standard matrix inversion techniques with runtime complexity ${\cal{O}}(P^3)$ for the matrix inversion and ${\cal{O}}(NP^2)$ for the matrix product.  This direct approach was the one advocated for encrypted regression by \citet{WuHaven12}, though dimensionality is constrained by school-book matrix inversion to enable homomorphic computation.

\subsection{Iterative methods}
\label{sec:LR_iterative}

We analyse two variants of an iterative algorithm, one which updates all parameters in $\bm{\beta}^{[k]}$ {\textit{simultaneously}} at each iteration $k$, using the vector $\bm{\beta}^{[k-1]}$ (\texttt{ELS-GD}); and another which updates these parameters {\textit{sequentially}}, using always the most current estimate for each parameter (\texttt{ELS-CD}).
%

The sequential update mode is related to the Gauss--Seidel method (coordinate descent), while the simultaneous update mode is related to the Jacobi method (gradient descent). See \citet[chapter 3]{Varga00}; \citet[chapter 7]{Bjorck96}.

As we will see below, there are two competing concerns here: optimisation efficiency and computational tractability within the cryptographic constraints, with these two types of updates having different properties.  We demonstrate that the properties of these methods in the encrypted domain means some standard optimality results no longer apply.

\subsubsection{Sequential updates via coordinate descent}
\label{sec:LR_CD}


Standard coordinate descent \citep[see e.g.][]{Wright15} for linear regression has the update:
\begin{equation}
\beta_{j_{k}}^{[k]} = \beta_{j_{k}}^{[k-1]} + \frac{\XjjT{j_{k}} (\y - \X \B^{\mathrm{old}})}{\XjjT{j_{k}} \Xjj{j_{k}}}
\end{equation}
where $\B^{\mathrm{old}}$ contains the components updated on the previous iteration or this one as appropriate, there being many variants for the schedule of coordinates to update.
However, this cannot be computed encrypted because the required data-dependent division is not feasible.
In an alternative variant we can replace this with a generic step-size $\delta$ and at each iteration $k$ choose one variable, say \mbox{$j_{k} \in \mathbb{N}_{1:P}$}, and update the corresponding regression parameter:
\begin{equation}\label{eq:cd}
\beta_{j_{k}}^{[k]}
= \beta_{j_{k}}^{[k-1]} + \delta \XjjT{j_{k}} (\y - \X \B^{\mathrm{old}})
\end{equation}
%
Two things are noteworthy:
first, these equations require only evaluation of polynomials; and
second, a rescaling is necessary since only integer polynomial functions can be computed homomorphically, and the parameter $\delta$ is usually not an integer.
We show how to perform this type of rescaling in the context of \S\ref{sec:LR_GD}.

\paragraph{Encrypted computation}
As each update uses the most recent estimate of every parameter, the MMD grows by $2$ with each parameter update (due to the term $\XjjT{j_{k}} \X \B^{\mathrm{old}}$).
This implies that for $K$ iterations over $P$ covariates the MMD is equal to $2KP$.
This renders the algorithm very expensive computationally, as it requires bootstrapping of ciphertexts for problems with even moderately large $P$.

\subsubsection{Simultaneous updates via gradient descent}
\label{sec:LR_GD}

In the case of gradient descent, for the objective function
$S(\B) = || \y - \X\B ||_{2}^{2}$
with
$\nabla S(\B)
={\partial S(\B)}/{\partial \B}
= -2\XT(\y - \X\B)$,
the update equations are:
%
%
\begin{align}\label{eq:LR_GD}
\B^{[k]}
&= \B^{[k-1]} - \delta \nabla S(\B^{[k-1]}) \\
&= \B^{[k-1]} + \delta \XT (\y - \X \B^{[k-1]}).
\end{align}
As with coordinate descent, these updates can be computed homomorphically, although the update equations must again be rescaled. This is strictly necessary to accommodate the transformed data, in order to overcome the fact that we cannot divide (see \S\ref{sec:LR_encoding}).  Letting $\delta \equiv 1/\nu$ for \mbox{$\nu \in \mathbb{N}$}, the transformed equations for the simultaneous updates are:
\begin{align}
\tilde{\B}^{[k]}
&\equiv 10^{\phi} \tilde{\nu} \tilde{\B}^{[k-1]} + \tilde{\bm{X}}^{T} (10^{k\phi}\tilde{\nu}^{k-1} \tilde{\y} - \tilde{\X} \tilde{\B}^{[k-1]}) \nonumber \\
&= 10^{(2k+1)\phi} \nu^{k} \B^{[k]} \label{eq:LR_GD_gd1t}
\end{align}
where now all transformed variables are represented with tildes, e.g., $\tilde{\X} = 10^{\phi} \X$ and similarly for the other variables, except the coefficients $\{\B^{[k]}\}_{k \in \mathbb{N}_{0:K}}$ as their scaling is iteration dependent (see supplementary materials, \S{1}).
The rescaling factors are independent of the data and known a priori, and so can be grouped during computation, e.g., $10^{k\phi} \tilde{\nu}^{k-1}$ in \eqref{eq:LR_GD_gd1t} can be encrypted as a single value.

Retrieval of the coefficients can be done by the secret key holder by computing
$\B^{[K]} \leftarrow \Dec(k_{s},\tilde{\B}^{[K]}) / (10^{(2K+1)\phi} \nu^{K})$.
Note the important difference between coordinate and gradient descent: for $K$ iterations, in CD each coefficient is updated $K/P$ times while in GD each coefficient is updated $K$ times.
%

\paragraph{Encrypted computation} There is a crucial difference between CD and GD in the encrypted domain:  GD reduces the multiplicative depth from $2KP$ to $2K$, which is independent of $P$, enabling scalability to higher dimensional models without bootstrapping or having to select parameters which support greater MMD.  As discussed in \S\ref{sec:HE}, bootstrapping in current FHE schemes is to be avoided wherever possible because it is very computationally expensive.  This is an interesting and important result, because it means that in the specific setting of encrypted computation, Nesterov's faster rates of convergence \citep{nesterov2012efficiency} compared to gradient descent in a randomised coordinate descent setting will not apply, as we will show.

For these computational reasons we focus primarily on gradient descent hereinafter.  For convergence in a regression setting recall:

\begin{lemma}[Convergence of \texttt{ELS-GD}]\label{thr:convergence}
Define
$\B^{[k+1]} \equiv \B^{[k]} + \delta \XT(\y - \X \B^{[k]})$
and let $\B^{[0]} \equiv \0{P}$.
Then,
$\lim_{k \to \infty} \B^{[k]} = (\XT\X)^{-1} \XT \y = \hat{\B}_{\text{ols}} $
for $\delta \in (0, 2/\mathcal{S}(\XT\X))$
where $\mathcal{S}(\XT\X)$ is the spectral radius of $\XT\X$.
\end{lemma}


The optimal choice of step size, in the sense that it minimises the spectral radius, is
$\delta_{\star} = 2/(\lambda_{\text{max}}+\lambda_{\text{min}})$,
implying an optimal spectral radius $\mathcal{S}_{\star} = (\lambda_{\text{max}}-\lambda_{\text{min}})/(\lambda_{\text{max}}+\lambda_{\text{min}})$, where $\lambda_{\text{max}}$ and $\lambda_{\text{min}}$ denote the largest and smallest eigenvalues of $\XT\X$, respectively \citep[see][Theorem 6.3 for all proofs]{RyabenkiiTsynkov06}.

\begin{lemma}[Oscillatory nature of \texttt{ELS-GD}]\label{thr:oscil}
The iterative process \eqref{eq:LR_GD} can be written as an oscillating sum:
\begin{equation}\label{eq:LR_oscilatoryGD}
\B^{[k]} = \sum_{n=1}^{k} (-1)^{n+1} \binom{k}{k-n} \delta^{n} (\XT\X)^{n-1} \XT\y
\end{equation}
\end{lemma}
This lemma is proved in the supplementary materials (\S{3}).
We show in \S\ref{sec:LR_vw} that it is possible to improve the convergence rate by using acceleration methods that exploit the oscillatory nature of the GD algorithm to accelerate the convergence of the series $\{ \B^{[k]} \}_{k \geq 0}$.


\subsection{Prediction}
\label{sec:LR_pred}

Note that the form of the GD equation \eqref{eq:LR_GD_gd1t} implies a common scaling factor, $10^{(2k+1)\phi} \nu^K$, for all parameters.
Performing encrypted prediction is then straightforward as it requires only the computation of the dot product $\tilde{y}_{i}
= \tilde{\X}_{i\sbl}^{T} \tilde{\B}^{[K]}
= 10^{(2K+1)\phi} \nu^K \hat{y}_{i}$,
where $\hat{y}_{i} = \X_{i\sbl}^{T} \B^{[K]}$ denotes the predicted values in the space of the original data and $\tilde{y}_{i}$ denotes the corresponding transformed version.
Upon decryption, rescaling can be done as before by the secret key holder. The procedure increases the MMD of the algorithm by 1.

The situation is more complex for coordinate descent since at the end of the final iteration each element of $\B^{[K]}$ will have different scaling.  Therefore the scaling must be unified before prediction, adding additional overhead.

\subsection{Inference}

Inference in the linear regression model (e.g., confidence intervals, hypothesis testing, variable selection) requires knowledge of the standard errors of the regression coefficients:
\begin{equation}
\mathbb{V}[\hat{\B}_{\text{ols}}] = \hat{\sigma}^{2} (\XT\X)^{-1}, \quad \hat{\sigma}^{2} = \bm{e}^{T} \bm{e} / (N-P).
\end{equation}
However, the matrix inversion is intractable under homomorphic computation except for very small $P$.
An alternative is to estimate the standard errors by bootstrapping the data and using the variability in the parameter estimates obtained.

\subsection{Regularisation}
\label{sec:LR_regularise}

$L_{2}$-regularised (ridge) least squares is easy to implement using the well known data augmentation procedure \citep{Allen74}:
\begin{equation}
\mathring{\X} =
\begin{bmatrix}
	\X  \\
	\sqrt{\alpha} \I{PP}
\end{bmatrix}
\qquad\text{and}\qquad
\mathring{\y} =
\begin{bmatrix}
	\y  \\
	\0{P}
\end{bmatrix}.
\end{equation}
OLS estimates when using the augmented data,
$(\mathring{\X},\mathring{\y})$,
are equivalent to RLS estimates when using the original data, $(\X,\y)$;
that is,
\begin{equation}\label{eq:LR_rls}
\hat{\B}_{\text{rls}}(\alpha)
=(\mathring{\X}^{T} \mathring{\X})^{-1} \mathring{\X}^{T} \mathring{\y}
= (\XT\X+\alpha \I{PP})^{-1} \XT\y.
\end{equation}

Because the augmentation terms $(\sqrt{\alpha} \I{PP}$ and $\0{P})$ are independent of the data $(\X$ and $\y)$, the iterative methods already developed for least squares (\S\ref{sec:LR_iterative}) can be used with the augmented data. 
Also note that the maximal eigenvalue is easily updated, $\mathring{\lambda}_{\text{max}} = \lambda_{\text{max}} + \alpha$
and so a new step size $\mathring{\delta}$ can be chosen without additional computation.


\subsection{Theoretical parameter requirements for the Fan and Vercauteren scheme}
\label{sec:LR_FVparams}

We provide results to guide the choice of cryptographic parameters for the encryption scheme of \citet{FanVercauteren12} --- hereinafter FV.  This is implemented in the \texttt{HomomorphicEncryption} R package \citep{AEH15review} and used in the examples.

FV represents data as a polynomial of fixed degree with coefficients in a finite integer ring.\footnote{The maximal degree and maximal ring element are tunable parameters, but even small increases make ciphertexts bigger and homomorphic operations slower.}  For example, $m$ is represented by $\mathring{m}(x) = \sum a_i x^i$ where $a_i$ are the binary decomposition of $m$, such that $\mathring{m}(2) = m$.  Addition and multiplication operations on the ciphertext result in polynomial addition and multiplication on this representation.

The transformed regression coefficients grow substantially during computation, so that we must ensure (i) the maximal degree of the FV message polynomial is large enough to decrypt $\tilde{\B}^{[K]}$;
and (ii) that the coefficient ring is large enough to accommodate the worst case growth in coefficients.
\begin{lemma}[FV parameter requirements for GD]
If data is represented in binary decomposed polynomial form, then after running the \texttt{ELS-GD} algorithm the degree and coefficient value of the encrypted regression coefficients is bound by:
\[\degr(\tilde{\B}^{[k]}) \leq \max \{4n+\degr(\tilde{\B}^{[k-1]}),(4k-1)n\}\] where $\degr({\B}^{[1]}) \leq 3n$ and $n \equiv (\phi+1)\log_{2}(10)$;
\begin{align*}
\mbox{and } ||\tilde{\B}^{[k]}||_{\infty} &\leq (4n+(n+1)^{2})NP || \tilde{\B}^{[k-1]} ||_{\infty} \\
& \qquad+ (4k-3)n(n+1)N
\end{align*}
where $||\tilde{\B}^{[1]}||_{\infty} \leq n(n+1)N$
\end{lemma}
This lemma is proved in the supplementary materials (\S{2}) and provides lower bounds on the choice of parameters $d$ and $t$ in the FV scheme.

Recall the MMD for GD is $2K$.  Theoretical bounds on security level \citep{LindnerPeikert11} and multiplicative depth \citep{LepointNaehrig14} --- together with the polynomial bounds and algorithmic MMD proved here --- then enable full selection of encryption parameters to guarantee security and computational correctness of the encrypted GD algorithm.

\section{Acceleration}
\label{sec:LR_accelerate}

Although \texttt{ELS-GD} is guaranteed to converge to the OLS solution, the rate of convergence can be slow, for instance when predictors are highly correlated.
Here we analyse some classic acceleration methods.

\subsection{Preconditioning}
\label{sec:LR_preconditioning}

\begin{figure}
\begin{center}
\includegraphics[trim={0 1.5cm 1cm 2.5cm},clip=true,width=0.5\columnwidth]{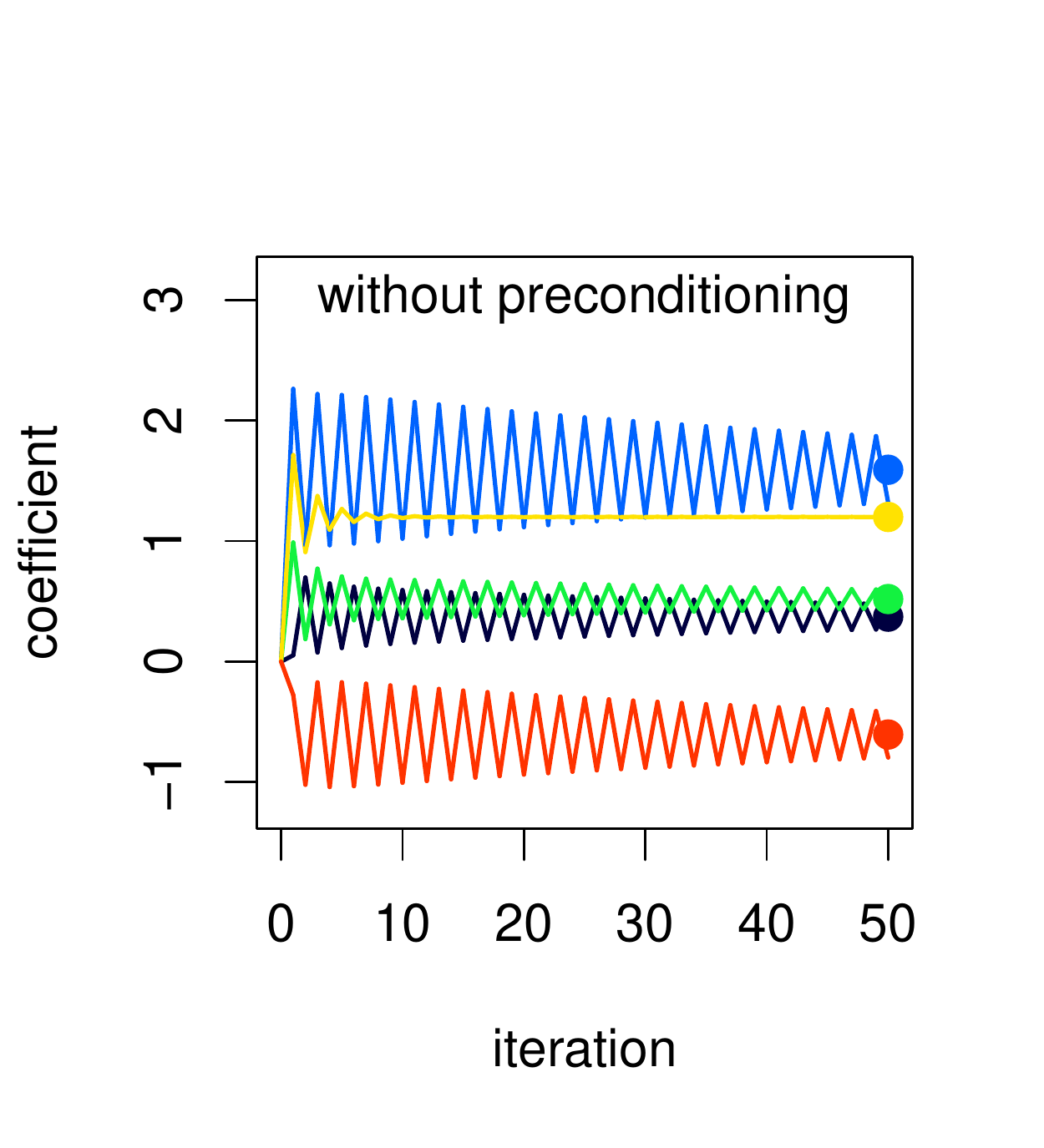}%
\includegraphics[trim={0 1.5cm 1cm 2.5cm},clip=true,width=0.5\columnwidth]{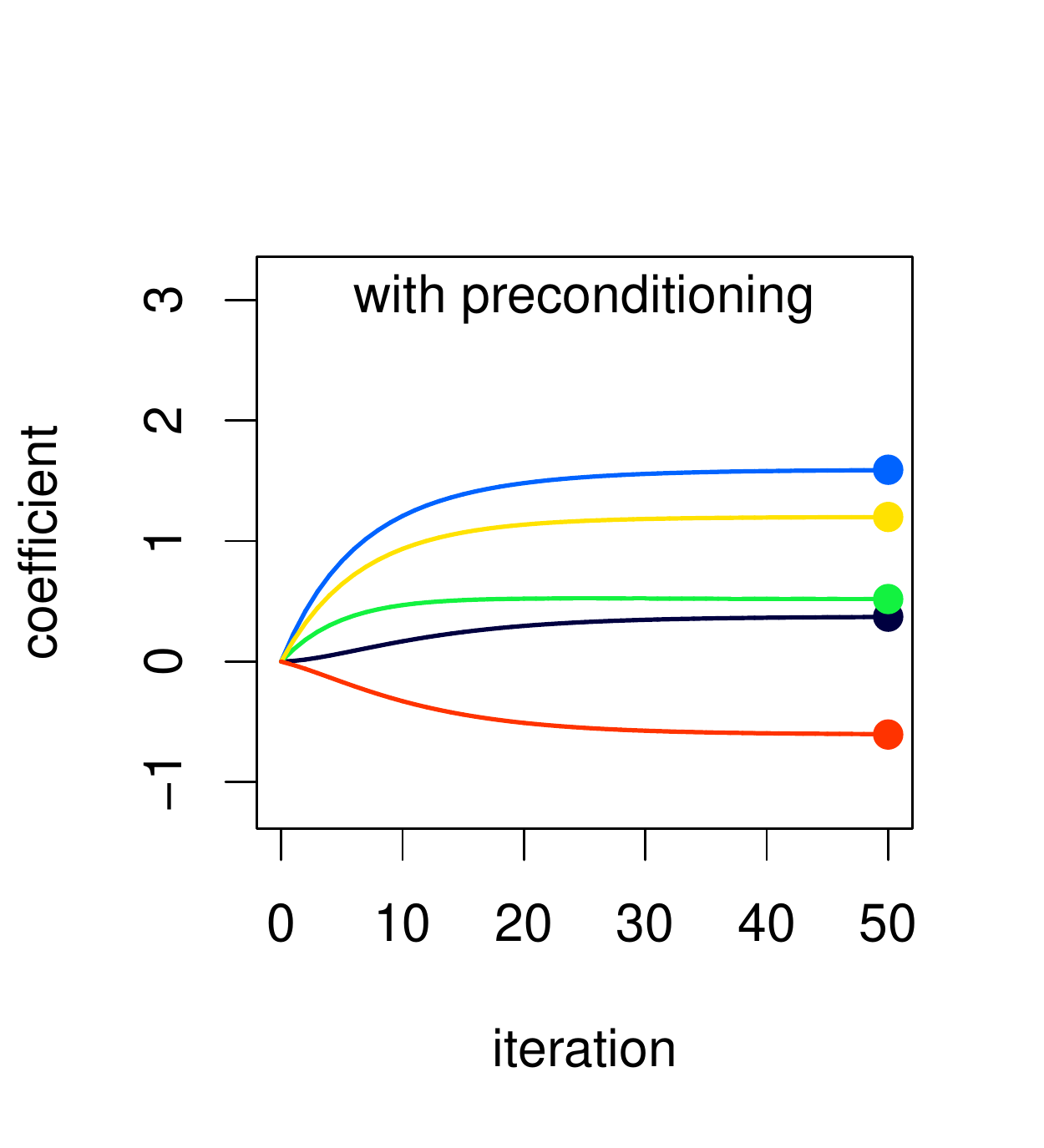}
\vspace{-20pt}
\caption{
The effect of diagonal scaling preconditioning on the convergence paths of \texttt{ELS-GD}. Full circles represent OLS estimates. [$N=100,P=5,\rho=0.1$]}
\label{fig:preconditioning}
\end{center}
\end{figure}

A preconditioning matrix $\bm{D}$ is often used to accelerate convergence of iterative methods \citep[chapter 7]{Bjorck96} by solving the preconditioned problem
\begin{equation}
\bm{D}^{-1} (\XT\X\B - \XT\y) = 0
\end{equation}
with the same solution as the original problem,
but having more favourable spectral properties.
A simple preconditioning is diagonal scaling of the columns of $\X$.
Let $\bm{D} = \text{diag}(d_{1}, \dots, d_{P})$ be a $P \times P$ diagonal matrix with diagonal entries $\{d_{1}, \dots, d_{P}\}$ where $d_{j} = || \Xj ||_{2}^{2}$.
Since $d_{j} \approx N$ for all $j \in \mathbb{N}_{1:P}$ as a result of standardisation (see \S\ref{sec:LR_encoding}) the preconditioning matrix becomes $\bm{D}^{-1} \approx N^{-1} \I{PP}$.
The preconditioned update equation is then
\begin{equation}
\B^{[k]} = \B^{[k-1]} + (\delta/N) \XT (\y - \X \B^{[k-1]})
\end{equation}
which differs from \eqref{eq:LR_GD} only in the step size.
Preconditioning smooths the convergence path, but the number of iterations required is still large (Figure~\ref{fig:preconditioning}).

\subsection{van Wijngaarden transformation}
\label{sec:LR_vw}

The \citet{vanWijngaarden65} transformation (VWT) is a variant of the Euler transform for accelerating the convergence of an alternating series.
Given the partial sums of an alternating series,
$S_{1,k} = \sum_{n=1}^{k} (-1)^{n} a_{n}$,
we can compute averages of these partial sums,
\begin{equation}\label{eq:vw_avg}
S_{z,k} = \frac{S_{z-1,k} + S_{z-1,k+1}}{2}, \quad z \in \mathbb{N}_{2:K}
\end{equation}
(and averages of the averages) to form a matrix of averaged partial sums.
For a finite number of terms ($K$), averages of partial sums are often closer to the limiting value, $S_{1,\infty}$, than  any single partial sum of the original series, $S_{1,k}$ ($k \in \mathbb{N}_{1:K}$), so that the alternating nature of the sequence is averaged out, damping its oscillatory behaviour and speeding-up convergence.

As shown in \eqref{eq:LR_oscilatoryGD}, the values $\{ \bm{\beta}^{[k]} \}_{k \in \mathbb{N}_{1:K}}$ computed with \texttt{ELS-GD} form an alternating series, making the VWT a candidate for accelerating the convergence to the true regression coefficients.

The implementation of \eqref{eq:vw_avg} has a simple, closed form solution.
For a stopping column $k^{\star} = \lfloor K/3 \rfloor + 1$ (as van Wijngaarden suggests) we compute the average partial sum
$S_{\star} = (2^{K-k^{\star}})^{-1} \sum_{n=k^{\star}}^{K} \binom{K-k^{\star}}{n-k^{\star}} S_{1,n}$,
and take this as our best approximation to the value of the series at convergence.
Notably, this can be implemented homomorphically with the exception of the division, but since the factor is independent of the data, we can compute instead $\tilde{S}_{\star} = 2^{K-k^{\star}} S_{\star}$ and incorporate the appropriate correction upon decryption; i.e., once \texttt{ELS-GD} is completed, the final VWT estimate is:
\begin{equation}
\tilde{\B}_{\text{vwt}} \leftarrow \sum_{k=k^{\star}}^{K} \binom{K-k^{\star}}{k-k^{\star}} \tilde{\B}^{[k]}.
\end{equation}
The computational cost of the VWT is minimal, involving approximately $2K/3$ additions and multiplications, and increasing the MMD by only 1.

\subsection{Nesterov's accelerated gradient}

Nesterov's accelerated gradient (NAG) achieves a convergence rate of \mbox{$\mathcal{O}(1/K^{2})$} as opposed to the $\mathcal{O}(1/K)$ achieved by regular GD \citep{Nesterov83}. The NAG algorithm can be written as follows:
\begin{subequations}\label{eq:nag}
\begin{align}
\bm{s}^{[k]}
&= \B^{[k-1]} + \delta \XT (\y - \X \tilde{\B}^{[k-1]}) \label{eq:nag1} \\
\B^{[k]}
&= \bm{s}^{[k]} + \eta_{k} (\bm{s}^{[k]} - \bm{s}^{[k-1]}), \qquad \eta_{k}<0. \label{eq:nag2}
\end{align}
\end{subequations}
The first step in NAG, \eqref{eq:nag1}, is the same as the standard GD step in \eqref{eq:LR_GD}. The extra step, \eqref{eq:nag2}, is proportional to the momentum term \mbox{$\bm{s}^{[k]} - \bm{s}^{[k-1]}$}, and is responsible for the acceleration.


These equations must also be rescaled for homomorphic computation (similarly to GD in \S\ref{sec:LR_GD}):
\begin{subequations}\label{eq:nagt}
\begin{align}
\tilde{\bm{s}}^{[k]}
&\equiv 10^{\phi} \tilde{\nu} \tilde{\B}^{[k-1]} + \tilde{\X}^{T} (10^{(2k-1)\phi} \tilde{\nu}^{k-1} \tilde{\y} - \tilde{\X} \tilde{\B}^{[k-1]}) \nonumber\\
&\equiv 10^{3k\phi} \nu^{k} \bm{s}^{[k]} \\
\tilde{\B}^{[k]}
&\equiv (10^{\phi} + \tilde{\eta}_{k}) \tilde{\bm{s}}^{[k]} - 10^{2\phi} \tilde{\nu} \tilde{\eta}_{k} \tilde{\bm{s}}^{[k-1]} \nonumber \\
&\equiv 10^{(3k+1)\phi} \nu^{k} \B^{[k]}
\end{align}
\end{subequations}
where $\tilde{\y} = 10^{\phi} \y$ is the transformed vector of responses, according to \S\ref{sec:LR_encoding}, and similarly for the remaining variables, except $\{ \tilde{\bm{s}}^{[k]}, \tilde{\B}^{[k]} \}_{k \in \mathbb{N}_{0:K}}$ which have iteration dependent scaling factors.

All scaling constants are independent of the data and known a priori, and so can be incorporated into the scaling by the secret key holder to obtain the final parameter estimates as: $\Dec(k_{s},\tilde{\B}^{[K]}) / (10^{(3K+1)\phi} \nu^{K})$.
Because of the extra acceleration step, \texttt{ELS-NAG} has a MMD equal to $3K$ (see Table~\ref{tab:LR_MMD}).
This is particularly interesting, because although Nesterov's method is state-of-the-art for unencrypted GD, the increase in MMD makes it costly for encrypted analysis.

\begin{table}
\captionsetup{width=\linewidth}
\caption{Maximum Multiplicative Depth (MMD).}
\label{tab:LR_MMD}
\vspace{-12pt}
\begin{center}
\begin{tabular*}{\linewidth}{lllc}
\hline
& \textbf{Algorithm} && \textbf{MMD}\\
\hline
& Preconditioned gradient descent && 2K \\
& van Wijngaarden transformation  && 2K+1 \\
& Nesterov's accelerated gradient && 3K \\
\hline
\end{tabular*}
\vspace{-10pt}
\end{center}
\end{table}

\section{Results}
\label{sec:LR_results}

In this section we empirically analyse the methods proposed for encrypted linear regression using simulated and real data (see supplementary materials, \S{4}, for details).
We use the implementation of the \cite{FanVercauteren12} cryptosystem provided by \citet*{AEH15review}.
The runtimes reported for encrypted analysis are on a 48-core server.
All error norms are root mean squared deviations w.r.t. OLS, and we use $\phi=2$ throughout.

\subsection{Simulations}
\label{sec:LR_simulations}

For simulations with \textit{independent data} we generate
\mbox{$\B \sim \mathcal{N}(\0{P},\I{PP})$},
\mbox{$\X \sim \mathcal{N}(\0{P},\bm{\Sigma})$} and
\mbox{$\y \sim \mathcal{N}(\X\B,\I{PP})$.}
For simulations with \textit{correlated data} we use Normal copulas and generate predictors whose pairwise correlations are all equal to $\rho$.

\begin{figure}[!ht]
\centering
\includegraphics[trim={0 0.8cm 1.5cm 3cm},clip=true,width=0.5\columnwidth]{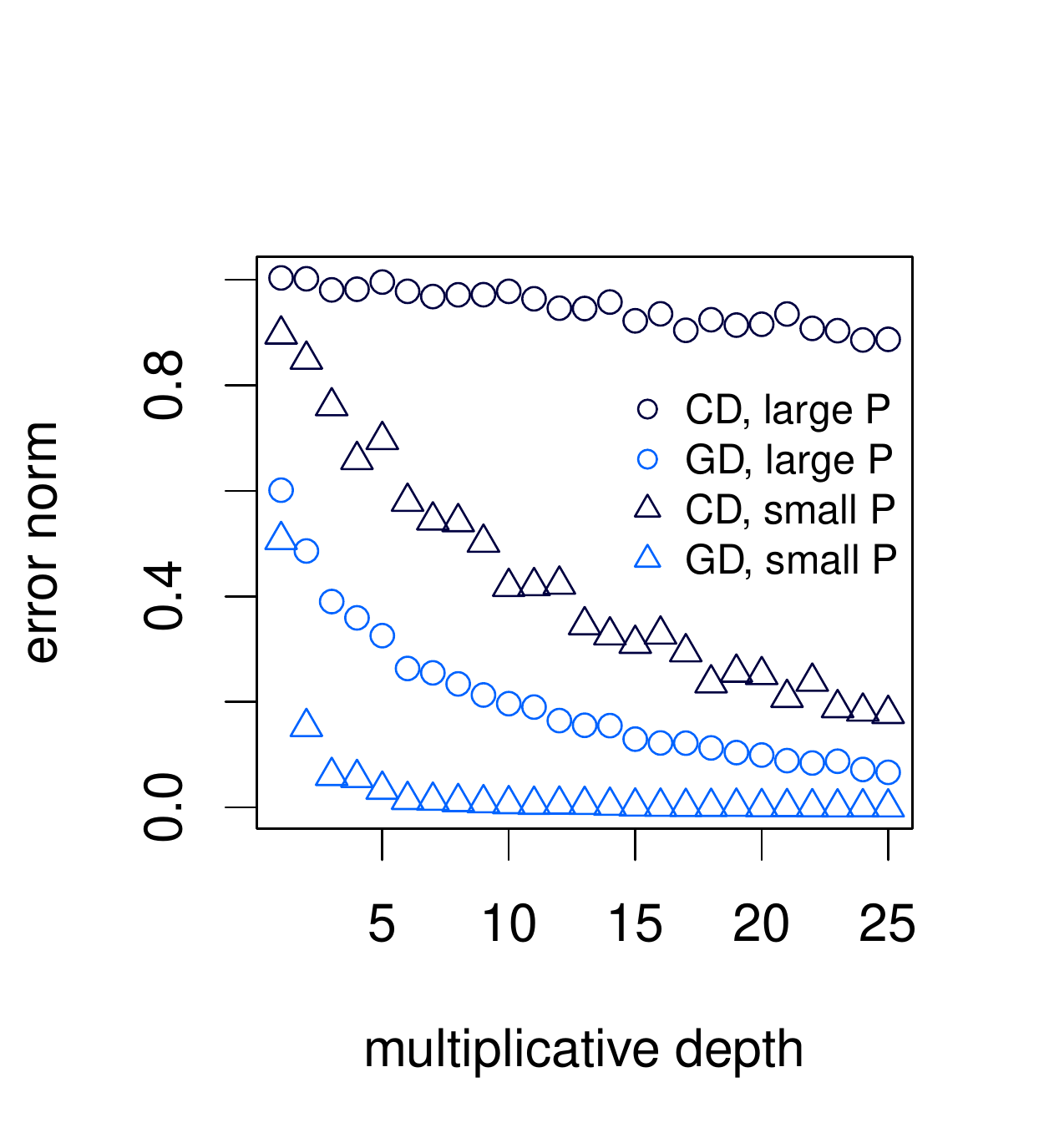}%
\includegraphics[trim={0 0.8cm 1.5cm 3cm},clip=true,width=0.5\columnwidth]{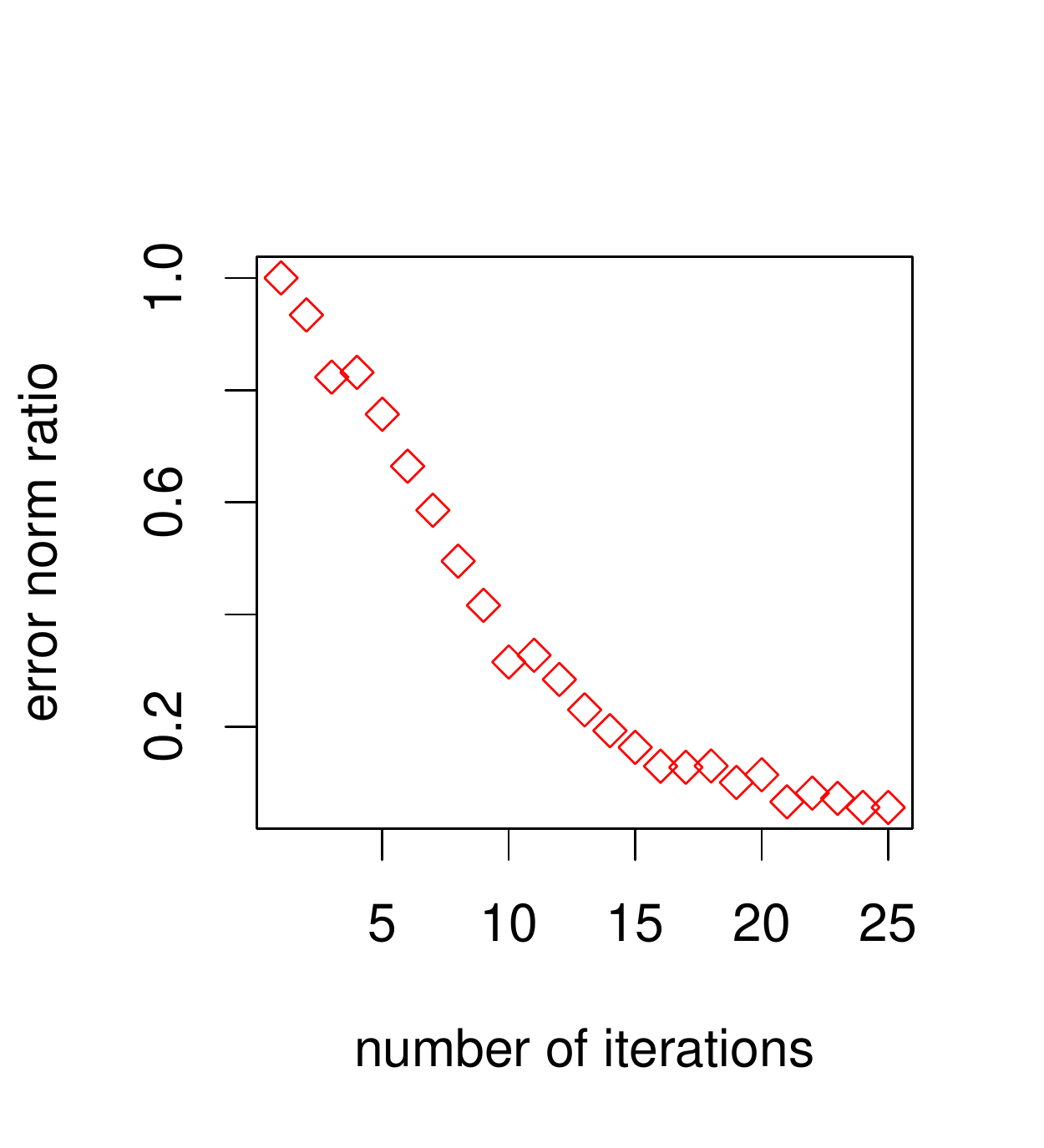}
\caption{
Error norms with respect to the OLS solution.
\textbf{[left]} 
\texttt{ELS-CD} vs. \texttt{ELS-GD}.
Comparison for fixed multiplicative depth
\textbf{[right]} 
acceleration via VWT. Ratios of error norm of \texttt{ELS-GD-VWT} to \texttt{ELS-GD}.
[$N=100$, large $P=50$, small $P=5$]
}
\label{fig:LR_CDvdGDvsVWT}
\includegraphics[trim={0 0.8cm 1.5cm 2cm},clip=true,width=0.5\columnwidth]{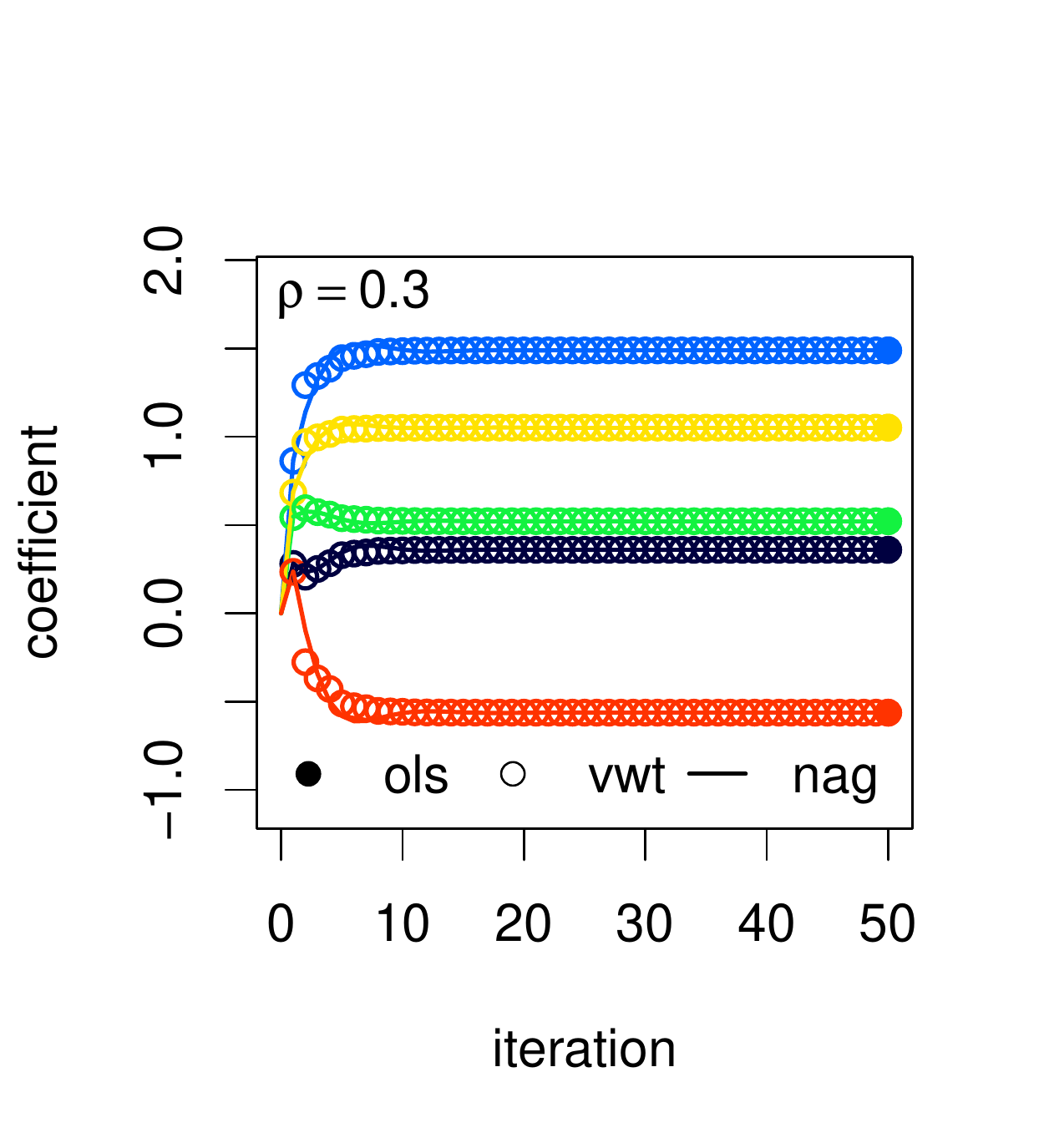}%
\includegraphics[trim={0 0.8cm 1.5cm 2cm},clip=true,width=0.5\columnwidth]{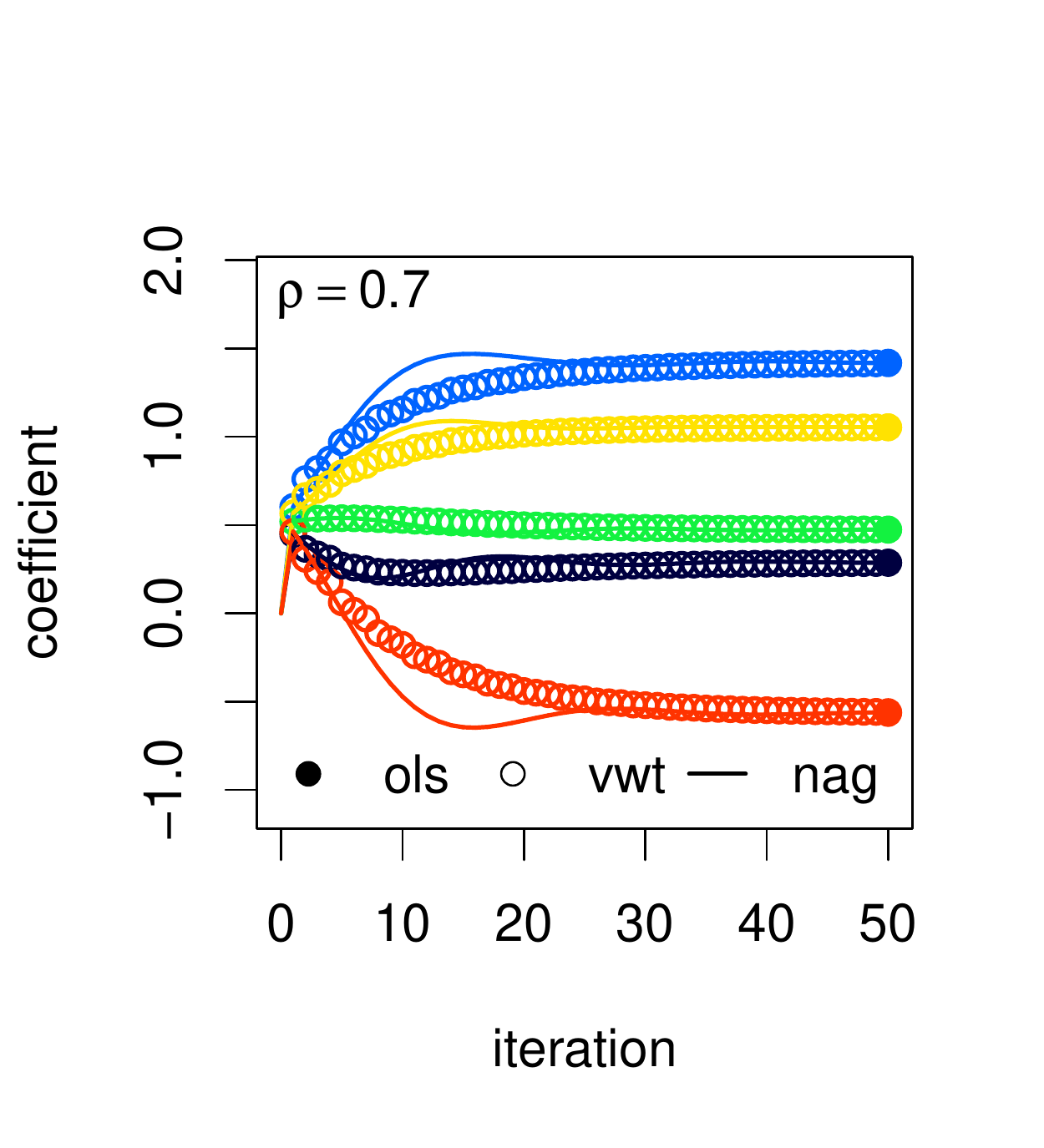}
\caption{
Convergence of \texttt{ELS-GD-VWT} and \texttt{ELS-NAG} for different levels of correlation: \textbf{[left]} $\rho=0.3$ \textbf{[right]} $\rho=0.7$.
[$N=100,P=5$]
}
\label{fig:LR_ConvM_P5}
\includegraphics[trim={0 0.8cm 1.5cm 2cm},clip=true,width=0.5\columnwidth]{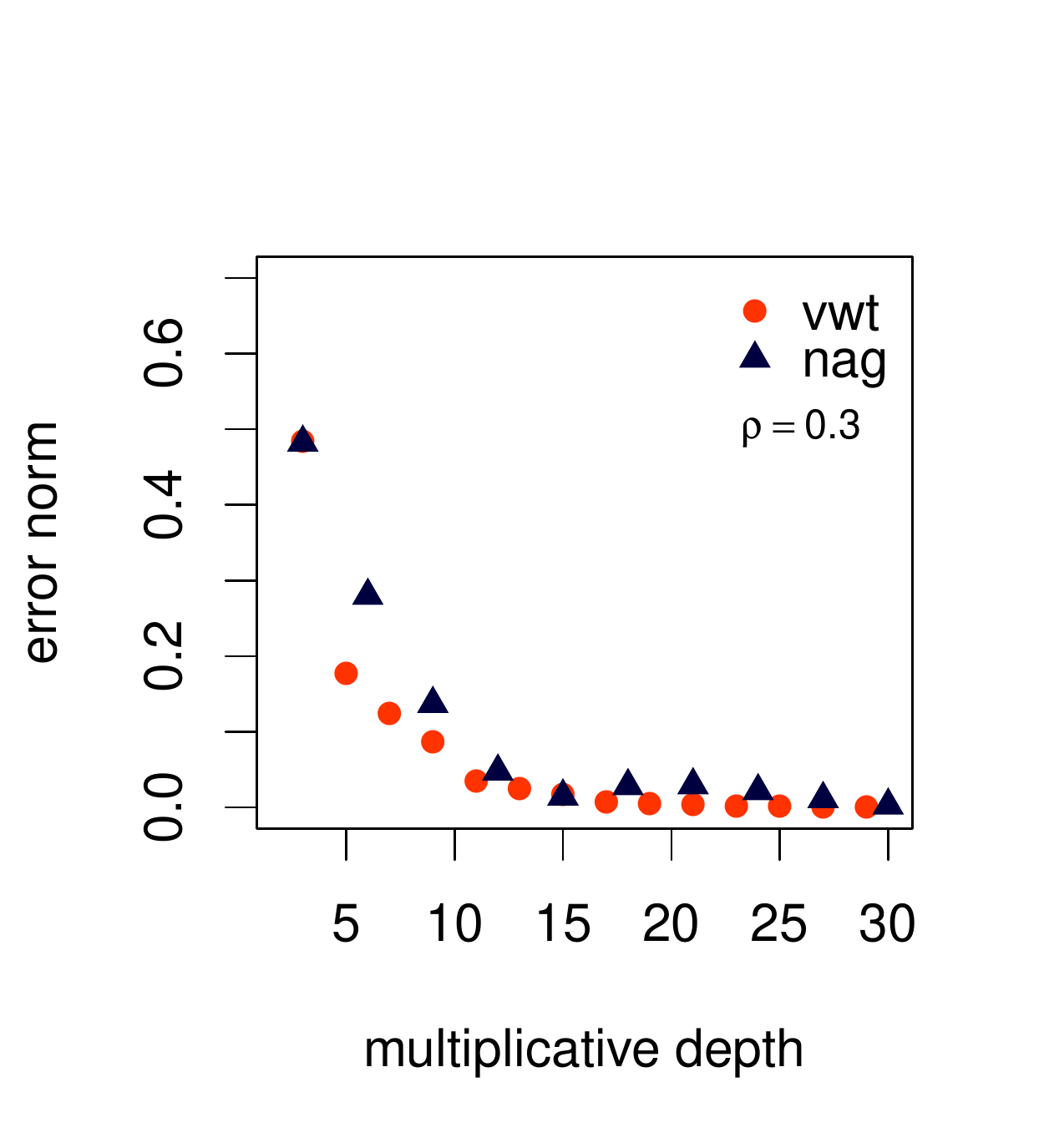}%
\includegraphics[trim={0 0.8cm 1.5cm 2cm},clip=true,width=0.5\columnwidth]{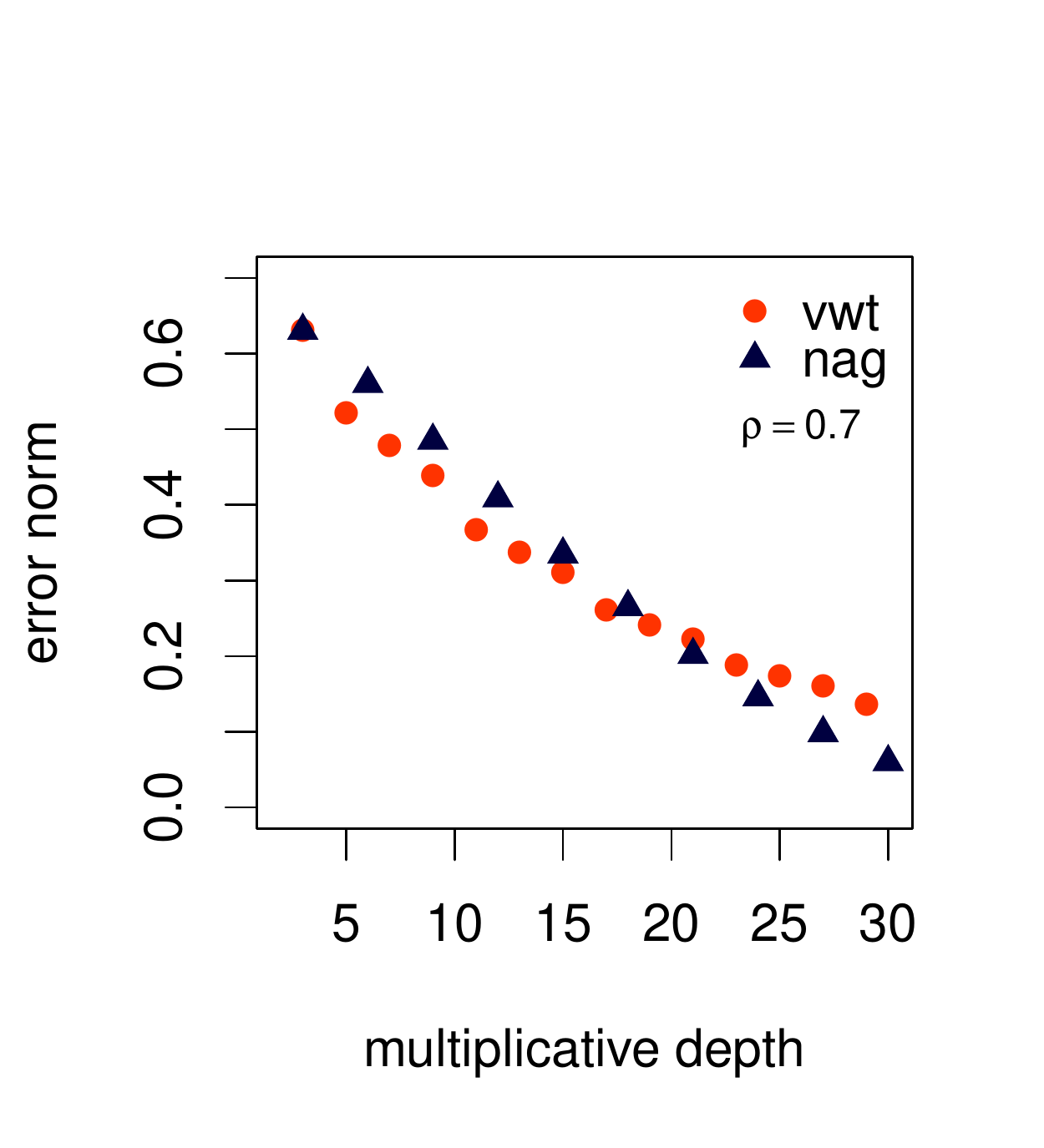}
\caption{
Error norm of \texttt{ELS-GD-VWT} and \texttt{ELS-NAG} as a function of multiplicative depth, for different levels of correlation: \textbf{[left]} $\rho=0.3$ \textbf{[right]} $\rho=0.7$. [$N=100,P=5$]
}
\vspace{-5pt}
\label{fig:LR_VWTvsNAG}
\end{figure}

Figure~\ref{fig:LR_CDvdGDvsVWT} (left) illustrates the computational properties of the coordinate and gradient descent methods for encrypted regression for a fixed MMD.  Since the MMD supported in the encryption scheme is a prime determinant of the computational cost of homomorphic operations, this serves as a proxy for the error as a function of encrypted computational complexity. \texttt{ELS-GD} clearly outperforms \texttt{ELS-CD} for a fixed encrypted computational cost, as expected from \S\ref{sec:LR_accelerate}.
Furthermore, Figure~\ref{fig:LR_CDvdGDvsVWT}, (right) shows the VWT provides additional acceleration in convergence relative to GD.

In general, higher correlation among predictors implies less favourable spectral properties for $\XT\X$, which in turn makes convergence slower for both \texttt{ELS-GD-VWT} and \texttt{ELS-NAG} (Figure~\ref{fig:LR_ConvM_P5}).
A fair comparison must control for the fact that the two algorithms have different encrypted computational complexities.
Using the MMD as a proxy again, \texttt{ELS-GD-VWT} typically outperforms \texttt{ELS-NAG} for fixed level of complexity  (Figure~\ref{fig:LR_VWTvsNAG}).
In very high correlation settings this relationship can be reversed, but only for large numbers of iterations, which it is desirable to avoid.

We stress that this choice is conditional on the encrypted computational framework considered here. It is particularly interesting that when working unencrypted, NAG is the state-of-the-art; but in the restricted framework of FHE, VWT empirically appears to be a better choice.

Convergence is affected by the number of predictors.
In particular, the maximum number of iterations required to reduce the norm of the initial error vector by a factor $e$ (reciprocal of the average convergence rate; \citealp[p.69]{Varga00}) gives us an idea of the relationship between number of predictors and speed of convergence.
For any level of correlation, this measure of complexity increases linearly with $P$ (see Figure~1 in the supplementary materials).

\begin{figure}[t]
\centering
\includegraphics[trim={0 1cm 1cm 1.5cm},clip=true,width=0.5\linewidth]{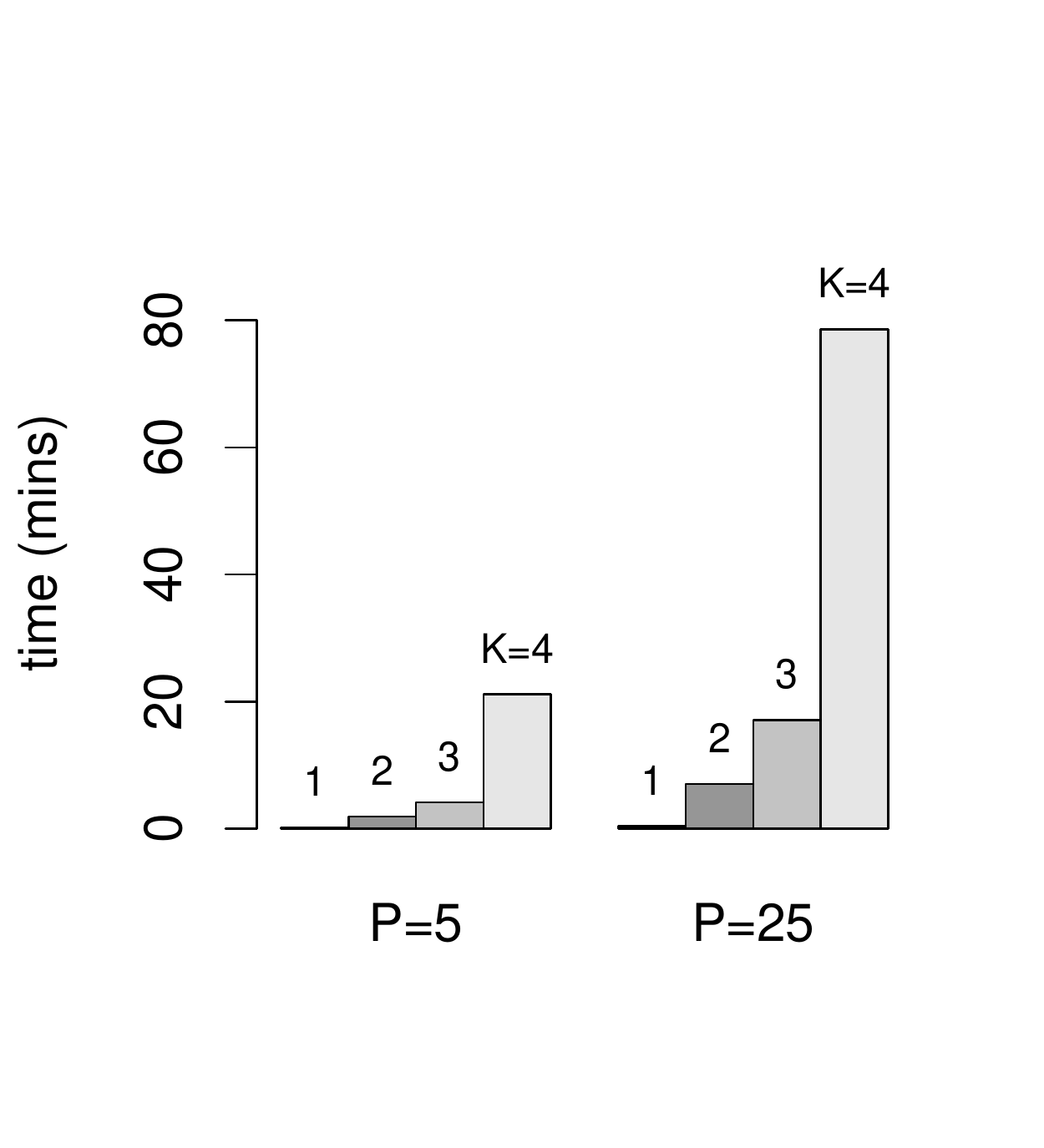}%
\includegraphics[trim={0 1cm 1cm 1.5cm},clip=true,width=0.5\linewidth]{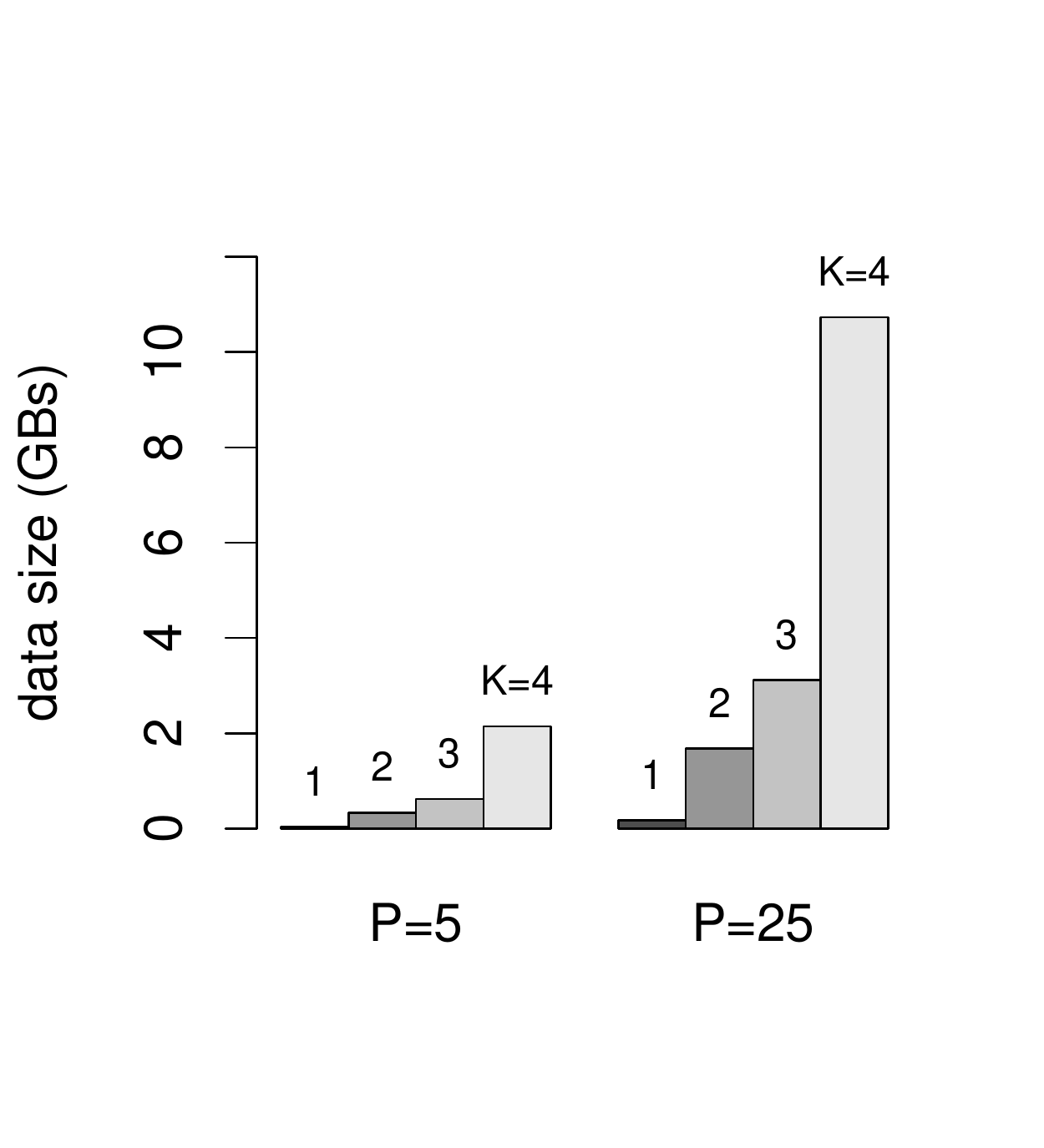}
\vspace{-25pt}
\caption{
Computational aspects of \texttt{ELS-GD} for different problem sizes $P\in\{2,25\}$, per 100 observations:
\textbf{[left]} runtimes (in minutes)
\textbf{[right]} encrypted data size in memory, $\{\X,\y\}$ (in gigabytes, excluding computational overheads).
}
\label{fig:LR_ComputationalAspectsSimul}
\end{figure}

Finally, the computational costs of \texttt{ELS-GD} are given in Figure~\ref{fig:LR_ComputationalAspectsSimul}.
Runtime grows quickly with the algorithm's multiplicative depth, which increases with the number of iterations.
However, for a fixed multiplicative depth the runtime is roughly linear in both $N$ and $P$.
Memory requirements grow in a similar fashion.

\subsection{Applications}
\label{sec:LR_realdata}

\paragraph{Mood stability data}

\begin{figure}[!ht]
\centering
\includegraphics[trim={0 0.8cm 1.5cm 2cm},clip=true,width=0.5\linewidth]{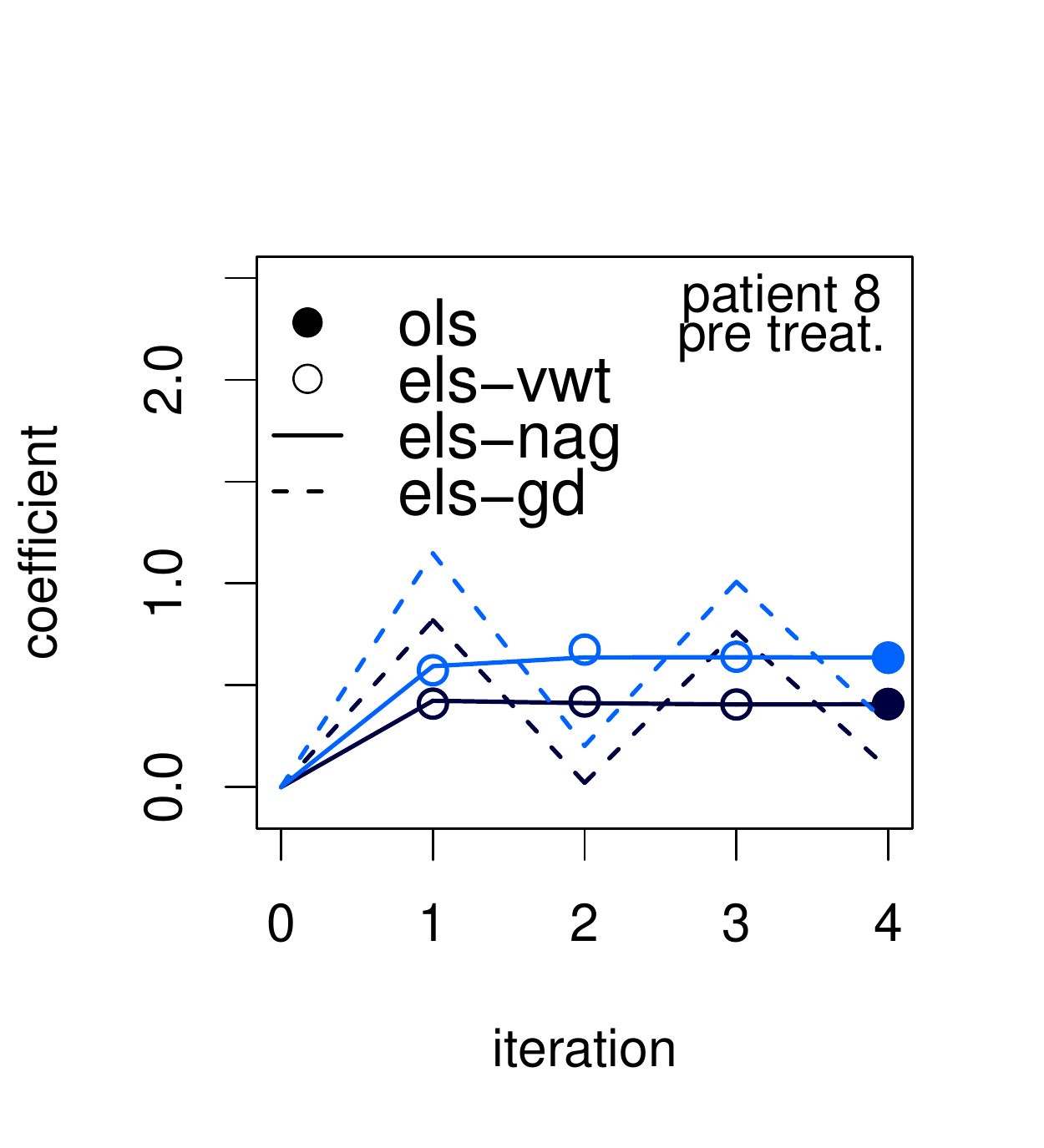}%
\includegraphics[trim={0 0.8cm 1.5cm 2cm},clip=true,width=0.5\linewidth]{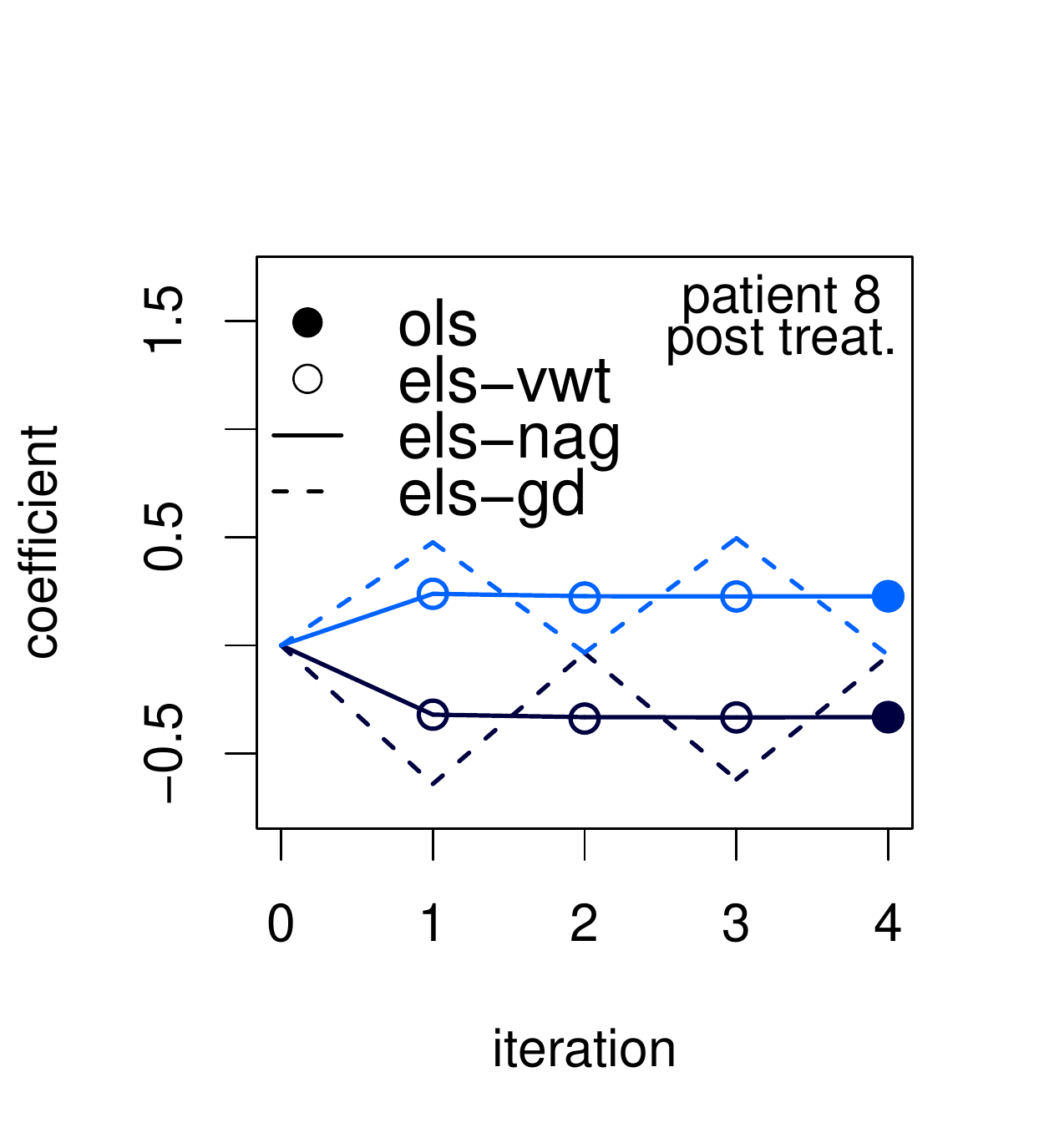}
\caption{
Convergence of different algorithms in the mood stability application (patient 8 shown): \textbf{[left]} pre treatment \textbf{[right]} post treatment. Lines of different colours represent different regression coefficients.
}
\label{fig:LR_bipolarFit}
\includegraphics[trim={0 0.8cm 1.5cm 2cm},clip=true,width=0.5\linewidth]{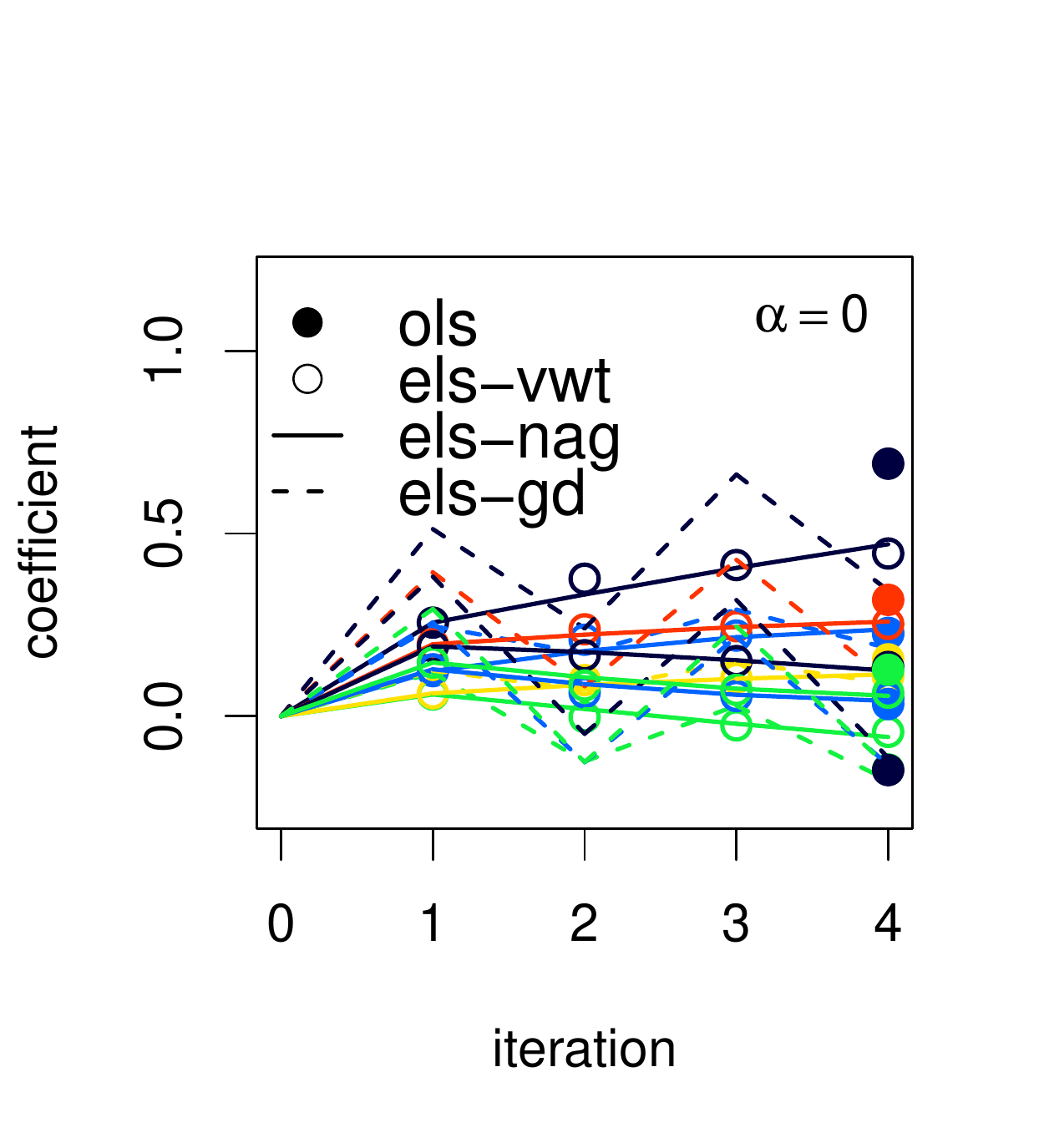}%
\includegraphics[trim={0 0.8cm 1.5cm 2cm},clip=true,width=0.5\linewidth]{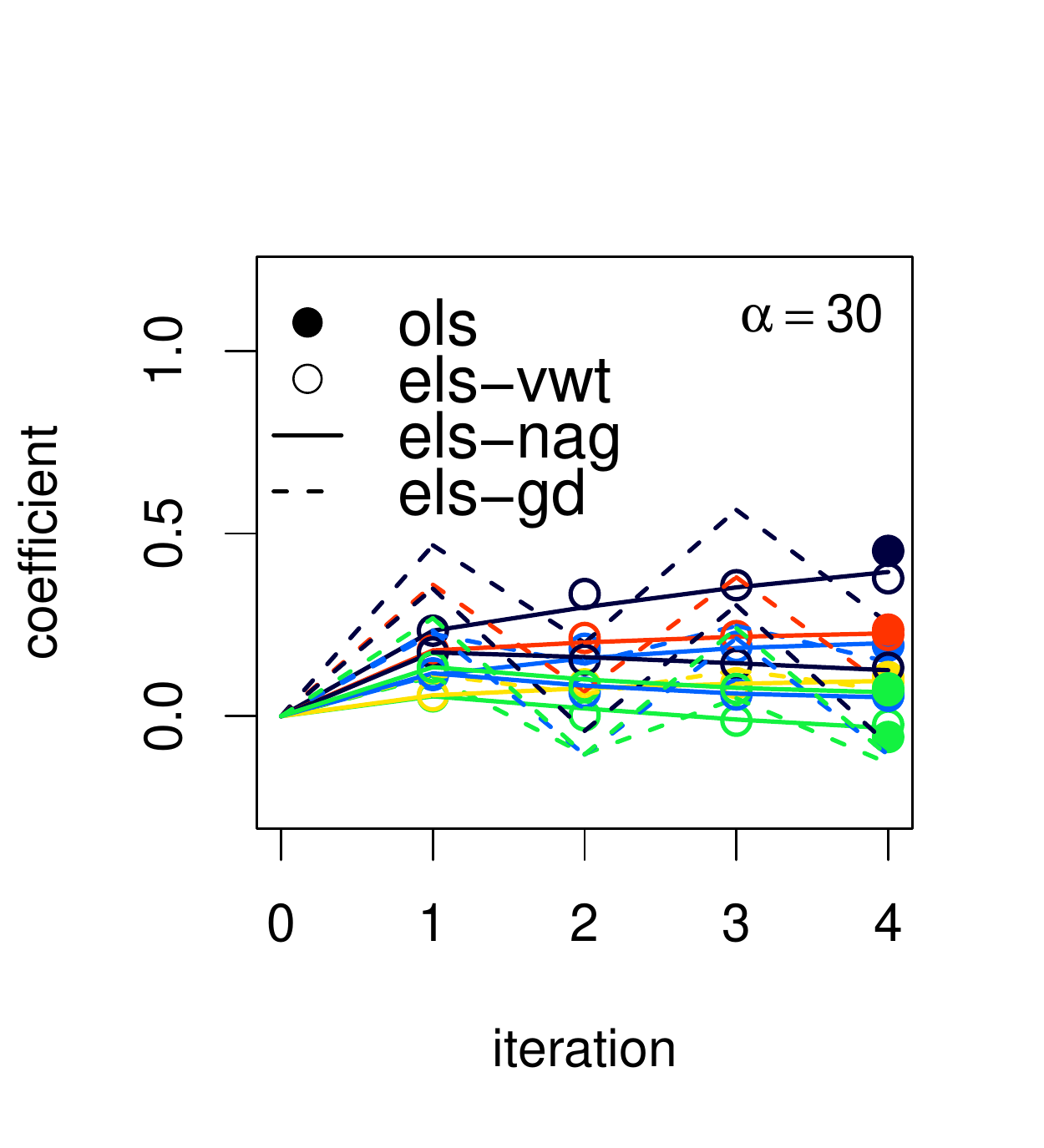}
\caption{
Convergence of different algorithms in the prostate data application: \textbf{[left]} without regularisation ($\alpha=0$)
\textbf{[right]} with regularisation ($\alpha=30$).
}
\label{fig:LR_prostateFit}
\includegraphics[trim={0.7cm 2cm 1.2cm 0},clip=true,width=\linewidth]{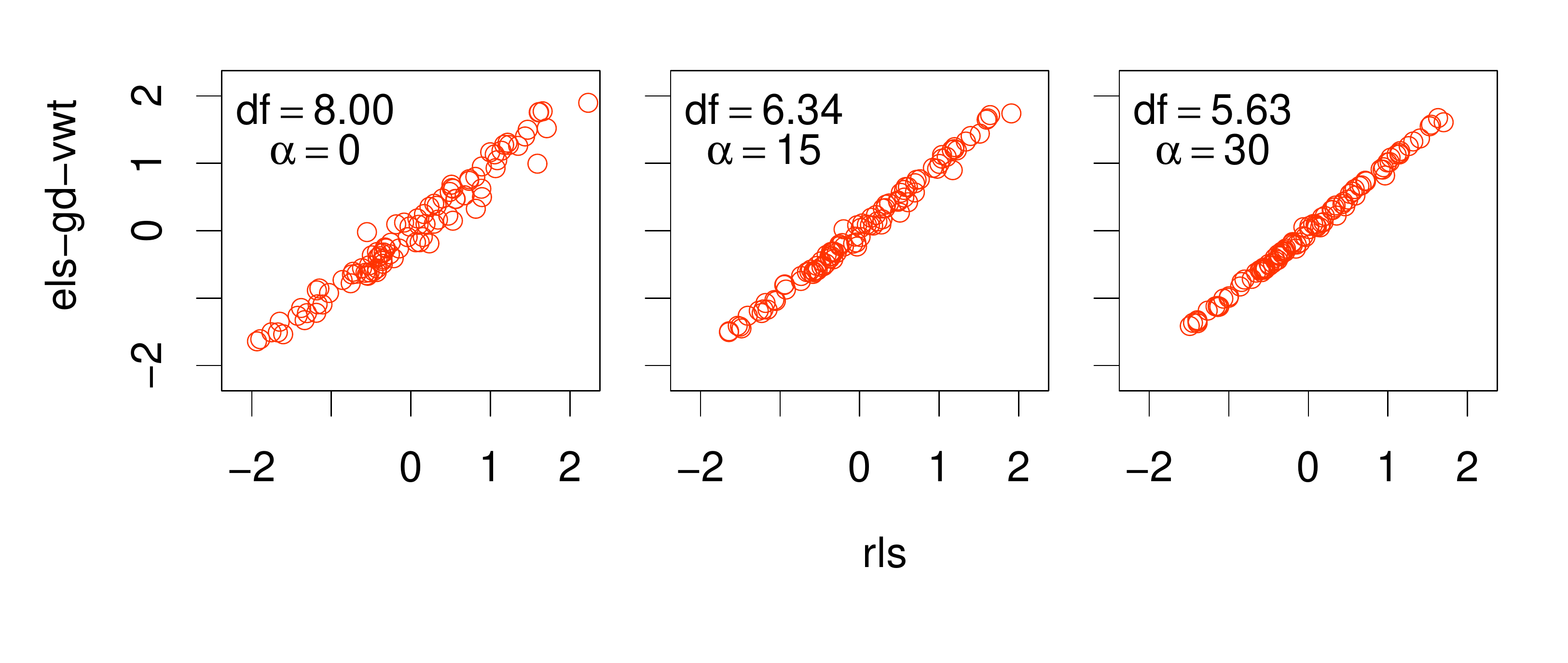}
\caption{
Predictions for the prostate dataset under different regularisation settings, $\alpha \in \{0,15,30\}$.
$\text{df} = \text{trace}(\X(\XT\X + \alpha \protect\I{PP})^{-1} \XT)$, effective degrees of freedom.
[$K=4$]
}
\label{fig:LR_predRLS}
\end{figure}

The first application is to mood stability in bipolar patients \citep{BonsallETAL12}.
Of interest in this application is the characterisation of the stochastic process governing the resulting time series, pre and post treatment, which we model as an autoregressive process of order two ($N=28,P=2$).
%
Convergence is achieved within 2 iterations ($||\B^{[2]}||_{\infty} \leq 0.04$; Figure~\ref{fig:LR_bipolarFit}).
The algorithm runs encrypted in 12 seconds and requires under 15 MBs of memory, excluding overheads.

\paragraph{Prostate cancer data}

The second application is to prostate cancer \citep{StameyETAL89}.
The model here is a standard linear regression ($N=97, P=8$).
Although not all parameters have completely converged by iteration 4 with unregularised \texttt{ELS-GD-VWT} ($||\B^{[4]}||_{\infty} \leq 0.26$; Figure~\ref{fig:LR_prostateFit}), the predictions are close to those produced by RLS (Figure~\ref{fig:LR_predRLS}).
The algorithm runs encrypted in 30 minutes and requires 3.5 GBs of memory ($K=4$).

For runtimes and memory requirements in these applications see Figure~2 in the supplementary materials.

\section{Discussion}
\label{sec:LR_discussion}

We demonstrated that in the restricted framework of FHE, traditional state-of-the-art methods can perform poorly.
Statistical and computational methods tailored for homomorphic computation are therefore required, which may differ from the state-of-the-art in an unrestricted framework.

For optimal convergence speed, the step size $\delta$ can be provided by the data holder, who can use the inequality
$\mathcal{S}(\XT\X) \leq ||(\XT\X)^{m}||^{1/m} \equiv B(m)$ to approximate $\mathcal{S}(\XT\X)$ to arbitrary precision, since $B(m) \to \mathcal{S}(\XT\X)$ as $m\to\infty$.

Choosing the penalty $\alpha$ is less straightforward.
Traditional methods involve cross-validation which is impossible under strict FHE.
Alternatively, it is possible to do rounds of communication and decryption between two parties to achieve this, in which case Differential Privacy can used as a way to guarantee security during the intermediate communication steps.


\subsubsection*{Acknowledgements}

P.M. Esperan\c{c}a: LSI-DTC doctoral studentship (EPSRC grant EP/F500394/1).
L.J.M. Aslett and C.C. Holmes: i-like project (EPSRC grant EP/K014463/1).

\bibliography{aistats}

\end{document}